\documentclass{article}

\PassOptionsToPackage{numbers, compress}{natbib}

\usepackage[preprint]{neurips_2026}

\usepackage[utf8]{inputenc} 
\usepackage[T1]{fontenc}    
\usepackage{hyperref}       
\usepackage{url}            
\usepackage{booktabs}
\usepackage{amsfonts}       
\usepackage{nicefrac}       
\usepackage{microtype}      
\usepackage{xcolor}         
\usepackage{amsmath}
\usepackage{xspace}

\usepackage{booktabs}       
\usepackage{graphicx}
\usepackage{multirow}
\usepackage{multicol}
\usepackage{color}
\usepackage{colortbl}
\definecolor{Gray}{gray}{0.86}
\usepackage[most]{tcolorbox}
\usepackage{tabularx}

\usepackage{amssymb}
\usepackage{bbding}
\usepackage{pifont}
\usepackage{wasysym}
\usepackage{amssymb}

\usepackage{algorithm}
\usepackage{algpseudocode}
\usepackage{amsmath, amssymb}
\usepackage{xcolor}
\usepackage{tikz}
\usetikzlibrary{tikzmark, calc, arrows.meta, fit, backgrounds}

\usepackage{subcaption}

\usepackage{svg}

\usepackage{pgfplots}

\usepackage{wrapfig}
\pgfplotsset{compat=1.18}

\usepackage{fontawesome}

\title{Ego Guidance: Enhancing Robot Policy Learning with Action-Free Human Egocentric Videos}

\vspace{-2cm}
\author{%
    \begin{minipage}[t]{\textwidth}
    \centering
        \textbf{Lin Long}$^{*1,5}$ \ \ \textbf{Tianren Ma}$^{*2,5}$ \ \ \textbf{Zhong Chu}$^{*3,5}$ \ \ 
        \textbf{Di Wu}$^{*4,5}$ \\
        \vspace{0.5em}
        \textbf{Maosong Cao}$^{5}$ \ \ 
        \textbf{Junbo Zhao}$^{1\dagger}$ \ \ 
        \textbf{Shanghang Zhang}$^{4\ddagger}$ \ \ 
    \end{minipage}
}

\begin{document}

\maketitle

\vspace{-0.9cm}
\begin{center}
     $^1$Zhejiang University \ $^2$University of Chinese Academy of Sciences \\
     $^3$Nanyang Technology University \ $^4$Peking University \ $^5$Ant Group \\
     \vspace{0.05cm}
     \small{$^*$Equal contribution \ $^{\dagger}$Project lead \ $^\ddagger$Corresponding author} \\
    \vspace{0.2cm}
     {\large\faGithub} \ \textcolor{cyan}{\url{https://llong-cs.github.io/EgoGuidance}}
\end{center}

\begin{abstract}
%


In this paper, we investigate how human egocentric videos can benefit robot policy learning without action supervision. 
Building on a pretrained VLM, we propose Shared Prediction on Action and Representation (SPAR), an MoT architecture in which future visual state prediction can provide independent and complementary learning signals alongside action prediction. 
%
%
We find that representations learned from egocentric videos improve downstream performance on out-of-distribution robot tasks, with further gains as video data scales. 
We also study the effects of objective choices and data curation, providing insights into how to leverage action-free videos for scalable robot learning.

\end{abstract}

\section{Introduction}

Embodied AI models are still in their infancy, eagerly absorbing every valuable source of world knowledge. Among these sources, human egocentric videos are particularly promising. Their large-scale availability, ease of acquisition, and close connection to the open-world environment make them an important potential source of data for embodied intelligence~\cite{graumanEgo4DWorld30002022a,lyuLDA1BScalingLatent2026,liOpenAoEOpenEgocentric2026, sunEgo4WAMWhatMatters2026}. However, precise action annotation is expensive to acquire, making most of the resources available only in video form (which we call \textit{action-free}). In this work, we investigate whether such action-free data can provide benefits to robot policy learning without action restoration.

Guided by first-principled considerations, rather than adopting a World Action Model (WAM) architecture~\cite{liangVideoGeneratorsAre2025,liLingbotVA2026,yeWorldActionModels2026a} that performs diffusion in a generative latent space, we start from a simpler and well-optimized Vision Language Model (VLM) backbone, expand it into a Mixture-of-Transformers (MoT), and jointly optimize the model with action and visual state prediction. 
Building on this \textbf{S}hared \textbf{P}rediction on \textbf{A}ction and \textbf{R}epresentation (\textbf{SPAR}), we investigate how different encoders for representing the predicted state affect performance.

We further conduct preliminary studies on several important factors, including data cleaning, diversity trade-off, and the scalability potential of pretraining with action-free egocentric videos. 

\section{Method}
\label{sec:method}

\begin{figure}[t]
  \centering
  \includegraphics[width=\linewidth]{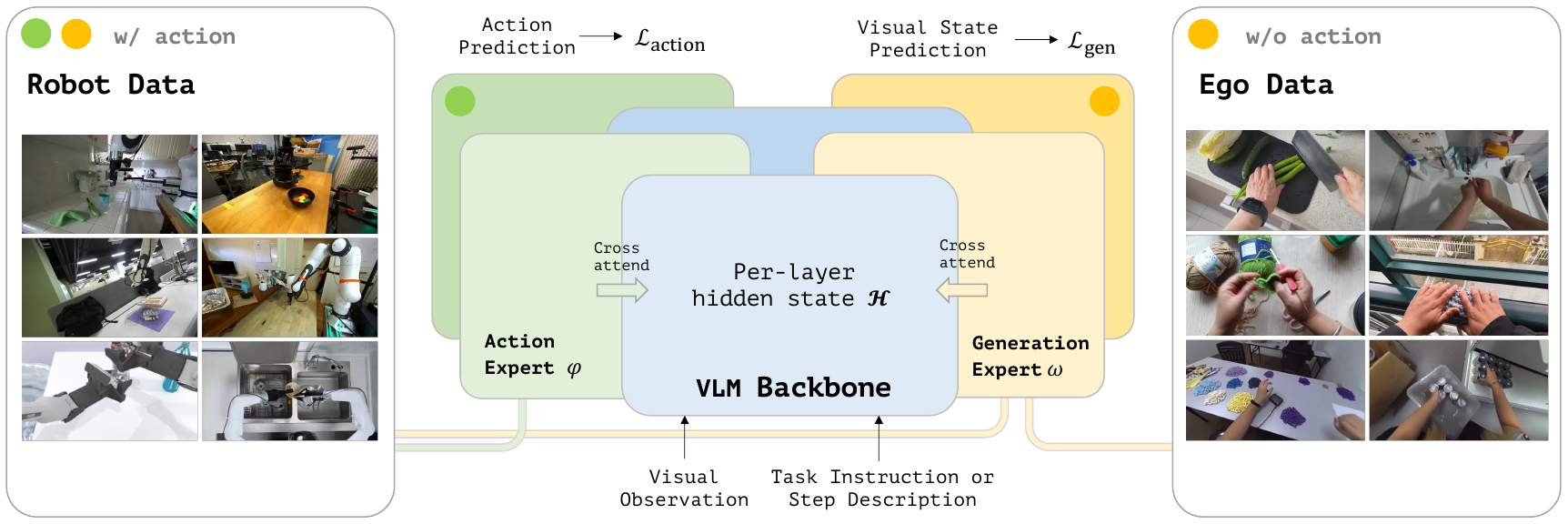}
  \caption{\textbf{Architecture Overview of SPAR.} Starting from a common VLA~\cite{community2026starvla,black2025pi05}, SPAR extends it with a generation expert that mirrors the action expert design, and predicts the representation of the future visual state simultaneously. Robot data supervises both experts, while the egocentric videos supervise only the generation expert. The VLM backbone can therefore be optimized with heterogeneous data composition. The generation expert is removed at inference.}
  \label{fig:arch}
\end{figure}

We adopt an MoT architecture~\citep{pi0,liang2024mot} that leverages a pretrained VLM~\cite{qwen35} to jointly model robot actions and future world states. 
As shown in Fig.~\ref{fig:arch}, two independently parameterized transformer stacks, \emph{i.e.,} an \emph{action expert} and a \emph{generation expert} are extended on the shared VLM backbone. 

\noindent\textbf{Action Expert. }Given the current observation and task instruction at time $t$ and horizon length $H$, the VLM's language model produces per-layer hidden states $\mathcal{H}=\{\mathbf{h}^{(l)}\}_{l=1}^{L}$. An action expert $f_\psi$ cross-attends to these states and predicts an action chunk $\mathbf{A}_t = a_{t:t+H}\in\mathbb{R}^{H\times d_a}$ with flow matching~\citep{lipman2023flow}: For a clean sample $\mathbf{x}$, Gaussian noise $\boldsymbol{\epsilon}\sim\mathcal{N}(\mathbf{0},\mathbf{I})$, and flow time $s\in[0,1]$, we define $\mathbf{x}^{s}=(1-s)\boldsymbol{\epsilon}+s\mathbf{x}$ and predict the velocity $\mathbf{x}-\boldsymbol{\epsilon}$. Applied to action chunks, this gives a common vision-language-action (VLA)~\citep{community2026starvla, bjorck2025gr00t} loss:
\begin{equation}
  \mathcal{L}_{\text{action}} = \mathbb{E}_{s,\boldsymbol{\epsilon}}\,\tfrac{1}{Hd_a}\big\|f_\psi(\mathbf{A}_t^{s},s;\mathcal{H})-(\mathbf{A}_t-\boldsymbol{\epsilon})\big\|^2_2.
  \label{eq:action_loss}
\end{equation}

\noindent\textbf{Generation Expert.}
Action labels provide a relatively low-dimensional, embodiment-specific training signal. To provide high-capacity and generalizable supervision for the shared VLM, we add a generation expert $g_\omega$ that predicts a future visual representation from the same hidden states $\mathcal{H}$ through layer-wise cross-attention. SPAR's two experts \textbf{do not} attend to each other, making the action and visual state prediction two independent proxy tasks for backbone optimization, \emph{i.e.,} the generation loss can improve the action pathway only by reshaping the shared hidden states.

Given the step description, the generation expert predicts the future visual state $\mathbf{Z}_t=\phi\big(\mathbf{o}_{t+\Delta}\big)\in\mathbb{R}^{N\times d}$,
that is, the features of the primary camera's future observation $\mathbf{o}_{t+\Delta}$ extracted by visual encoder $\phi$. Here, we align $\mathbf{o}_{t+\Delta}$ with $a_{t+H}$ to ensure that they are sampled in the same timestep.

The prediction of $\mathbf{Z}_t$ can be supervised by either flow matching or regression: For the \textit{flow-matching} version, we apply a similar velocity target used for actions to the future visual features, with noising schedules modified following RAE~\cite{zheng2025rae}. The noised target $\mathbf{Z}_t^{s}$ is linearly projected to the expert's width and given learned positional embeddings, the blocks are conditioned on $s$ through AdaLN, and an output projection predicts the velocity, optimized by $\mathcal{L}^{\text{flow}}_{\text{gen}}$. For the \textit{regression} version, we set the timestep to constant $s=0$, and use a three-layer MLP to map the expert's output to the prediction. The expert is therefore optimized by $\mathcal{L}^{\text{reg}}_{\text{gen}}$ to directly predict the standardized target $\bar{\mathbf{Z}}_t=(\mathbf{Z}_t-\mu_\phi)/\sigma_\phi$, where $\mu_\phi$ and $\sigma_\phi$ are scalar statistics of encoder $\phi$ (App.~\ref{app:impl}).

The losses are written as follows, and can both be denoted by $\mathcal{L}_{\text{gen}}$:
\begin{equation}
  \mathcal{L}^{\text{flow}}_{\text{gen}} = \mathbb{E}_{s,\boldsymbol{\epsilon},t}\,\tfrac{1}{Nd}\big\|g_\omega(\mathbf{Z}_t^{s},s;\mathcal{H})-(\mathbf{Z}_t-\boldsymbol{\epsilon})\big\|^2_2, \ \ \ \ \mathcal{L}^{\text{reg}}_{\text{gen}}=\mathbb{E}_{t}\,\tfrac{1}{Nd}\big\|g_\omega(\mathcal{H})-\bar{\mathbf{Z}}_t\big\|^2_2.
  \label{eq:diff_loss}
\end{equation}

\noindent\textbf{Co-training with Action-free Videos.}
Each data source supervises the losses it can provide:
\begin{equation}
  \mathcal{L}_{\text{robot}}=\mathcal{L}_{\text{action}}+\lambda\,\mathcal{L}_{\text{gen}},\qquad
  \mathcal{L}_{\text{ego}}=\lambda\,\mathcal{L}_{\text{gen}},
  \label{eq:source_losses}
\end{equation}
with $\lambda=1$. Each training step combines a robot and an egocentric mini-batch. Both losses back-propagate through the cross-attention into the LLM, while the action expert is updated only by robot data. The effect of egocentric videos on the policy is therefore entirely mediated by the backbone representation they help shape.

\noindent\textbf{Inference.}
At inference, the generation expert is removed following Fast-WAM~\cite{yuanFastWAMWorldAction2026}, and actions are obtained by integrating the action expert's velocity field from Gaussian noise with a few Euler steps. The deployed policy thus has the same architecture and cost as a standard VLA. All reported numbers are averaged across three runs with different random seeds.

\section{Experiment}
We describe the datasets and training settings in this section. See Appendix~\ref{app:impl} for more details.

\subsection{Datasets}
\paragraph{Robot Data. }
We curate the dataset from DROID~\cite{khazatskyDROIDLargeScaleInTheWild2024} by retaining successful episodes and filtering out episodes with no motion or without valid instruction or action annotations. This yields approximately 38,000 episodes totaling 230 hours of robot data (66\% retention), which we co-train with egocentric video data. For downstream fine-tuning, we use LIBERO~\cite{liu2023libero} demonstrations following the standard protocol~\cite{fei25libero-plus} for simulated evaluation. For the final real robot scenario, we extend the robot data with more datasets~\cite{liu2023libero,chen2025robotwin2,contributors2024agibotworldrepo,internroboticsrepo,abc2026,simpleai2026hifiumi}, totaling about $8,000$ hours, and internal demonstrations collected via tele-operation for downstream fine-tuning. 

\paragraph{Egocentric Data. }
We use Ego4D~\cite{graumanEgo4DWorld30002022a} for our preliminary experiments. As it does not provide usable action labels, we treat it as action-free egocentric video data. Its raw videos comprise 3,976 hours, and we describe our curation in Section~\ref{sec:ego_data}. To increase scale, we also use Open-AoE~\cite{liOpenAoEOpenEgocentric2026}, which provides approximately 1M annotated video clip--caption pairs on 8,000 tasks, spanning 2,000 hours. For Open-AoE, we use the released clips and language descriptions following the dataset's original curation, without applying the Ego4D-specific gates.

\subsection{Training}
We compared two training strategies in the primary experiments. In single-stage training, we jointly train on the robot demonstrations that are aligned with the downstream evaluation tasks and egocentric videos. In two-stage training, we first co-train on a broader collection of robot demonstrations and egocentric videos, then fine-tune on downstream-aligned demonstrations alone. The single-stage strategy consistently yields poorer generalization, with little variation across experimental settings. This behavior suggests overfitting to the limited robot data and provides little insight into the effect of egocentric pretraining. We therefore use the two-stage strategy for SPAR as default.

\paragraph{Pretraining.}
We co-train on robot and egocentric data. Both robot and egocentric data supervise the prediction on future visual-state representation. Robot data also supervise the prediction of the subsequent action chunk as common VLA models. 

\paragraph{Downstream Fine-tuning.}
We then fine-tune on robot demonstrations with $\mathcal{L}_{\text{robot}}$. Because pretraining is intended to shape the shared VLM representation, the action expert can either be reinitialized for the downstream task or trained further from its pretrained weights. In our experiments, reinitializing the action expert yields a modest improvement.

\subsection{Preliminary Findings}

\begin{figure}[t]
  \centering
  \includegraphics[width=\linewidth]{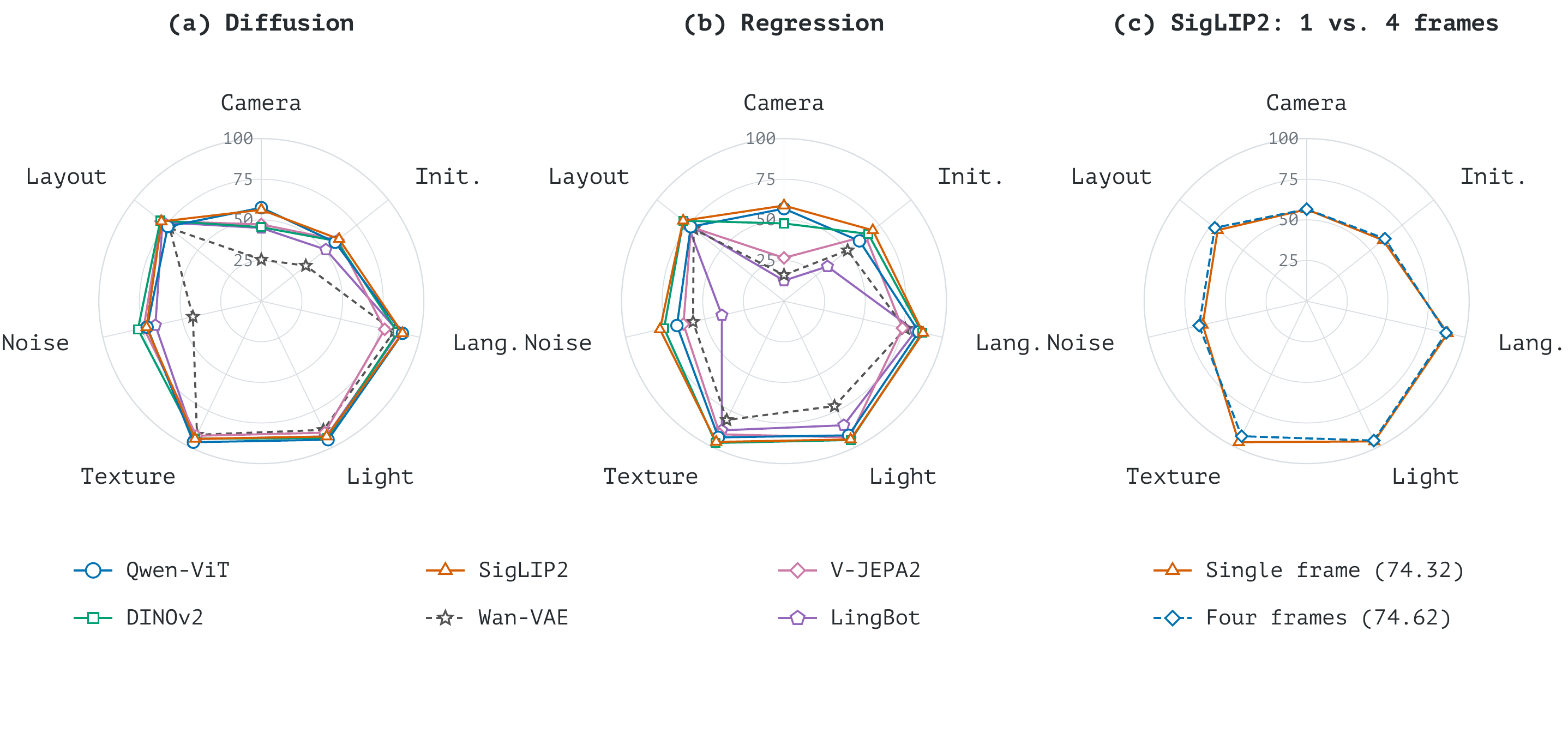}
  \caption{\textbf{Success rates across seven distribution shifts.} (a) Diffusion and (b) regression prediction of one future state frame with six visual targets; the two panels share a target legend. (c) Diffusion prediction of one versus four future frames with SigLIP2 in a separate frame-count experiment.}
  \label{fig:main_results_radars}
\end{figure}

\paragraph{Representational target helps generalization.}
Figure~\ref{fig:main_results_radars}(a) compares the per-dimension success rates on Libero-Plus across states encoded by six different encoders~\cite{qwen35,Siglip2025,assran2025vjepa2,DINOv2,wan2025,lingbotvision}. The target representation substantially affects performance under distribution shift. SigLIP2 achieves the highest reported average success rate (76.2\%), followed by Qwen-ViT (75.6\%) and DINOv2 (74.9\%). All five representation targets outperform the reconstruction-oriented Wan-VAE latent (60.0\%) in this comparison, consistent with the view that predictive supervision in a learned visual feature space is more useful for policy learning than reconstructive latents. No target dominates every shift: DINOv2 obtains the highest diffusion scores for sensor noise and object layout, whereas Qwen-ViT leads under lighting and background-texture changes. The choice of visual target affects both overall performance and the types of perturbations to which the policy generalizes.

\paragraph{Regression supervision also works.}
Interestingly, figure~\ref{fig:main_results_radars}(b) shows that direct regression of future visual features also provides useful supervision. Its per-dimension results are broadly comparable to diffusion for several targets, particularly SigLIP2 and DINOv2, suggesting that the benefits of visual prediction in our model do not depend on a diffusion objective.

\paragraph{State prediction can be single framed.}
During training, the timestep $t$ is already randomly sampled and reflects the request for dynamic modeling. We then investigate whether predicting a single frame works, as it significantly reduce the training cost. For the high-dimensional SigLIP2 target (256 tokens of width 1,152; App.~\ref{app:impl}), Fig.~\ref{fig:main_results_radars}(c) compares predicting one future frame with predicting four using the same diffusion objective. Their reported average success rates are 74.32\% and 74.62\%, respectively, a difference of only 0.30 percentage points. Together with the target comparison, we set the default design as predicting a single future frame in SigLIP2's visual representation space.

\section{What Matters in Action-Free Egocentric Videos?}
\label{sec:ego_data}

\begin{figure}[t]
    \centering
    \includegraphics[width=\linewidth]{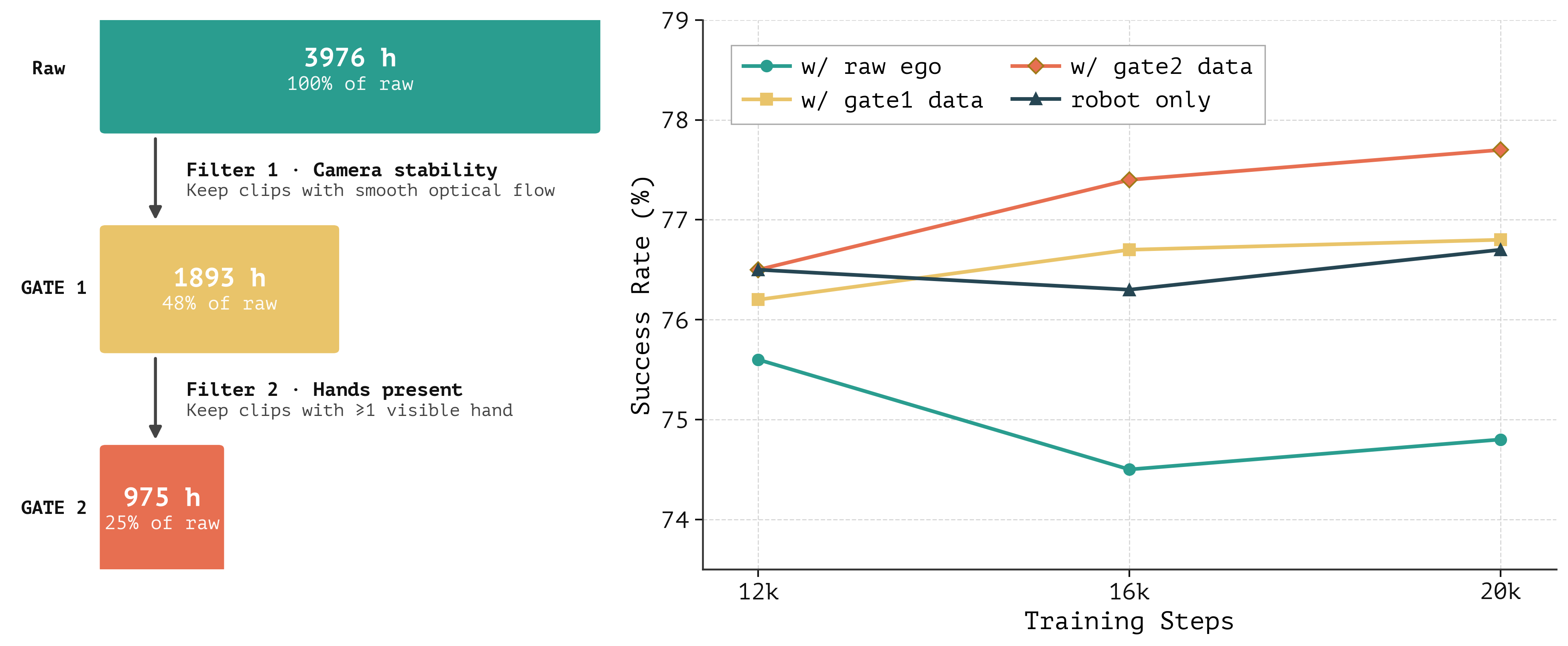}
    \caption{\textbf{Ego4D curation and corresponding OOD task success rates.} Left: hours retained after the motion-quality and visible-hand gates. Right: robot-only training compared with co-training on raw Ego4D, Gate-1 clips, or clips passing both gates, across fine-tuning steps.}
    \label{fig:ego4d_curation}
\end{figure}

\paragraph{Data Processing.}
We divide Ego4D videos into uniform 5-second clips and apply two cumulative quality gates based on EgoSmith~\cite{zhong2026EgosteerorEgosmith}. Gate 1 uses corner and optical-flow checks to retain clips with limited camera motion and reliable flow estimates; Gate 2 retains clips with at least one sufficiently visible hand in a prescribed image region. The first gate retains approximately 1,893 hours (48\% of the raw duration), and both gates together retain 975 hours (approximately 25\%; Fig.~\ref{fig:ego4d_curation}). We tested further fine-grained EgoSmith-based quality filters but observed no additional benefit, so they are not part of the final pipeline. We then use a locally deployed Qwen3.6-35B-A3B model to generate a short English step description from 15 frames sampled at 10-frame intervals per clip. This description serves as the language instruction for future-state prediction. Appendix~\ref{app:ego4d_curation} gives the exact gate criteria and captioning instructions.

The curation is consequential for downstream learning (Fig.~\ref{fig:ego4d_curation}). Training with raw Ego4D clips underperforms the robot-only baseline, whereas the clips retained after both gates surpass it at 16k and 20k fine-tuning steps. Gate 1 alone provides at most a small improvement over robot-only training; requiring visible hands in addition to stable, reliable optical flow gives the strongest result. Thus, selecting clips with visible hand activity makes action-free video a more useful source of supervision than simply adding unfiltered video.

\begin{figure}[t]
  \centering
  \begin{subfigure}[b]{0.47\linewidth}
    \centering
    \includegraphics[width=\linewidth]{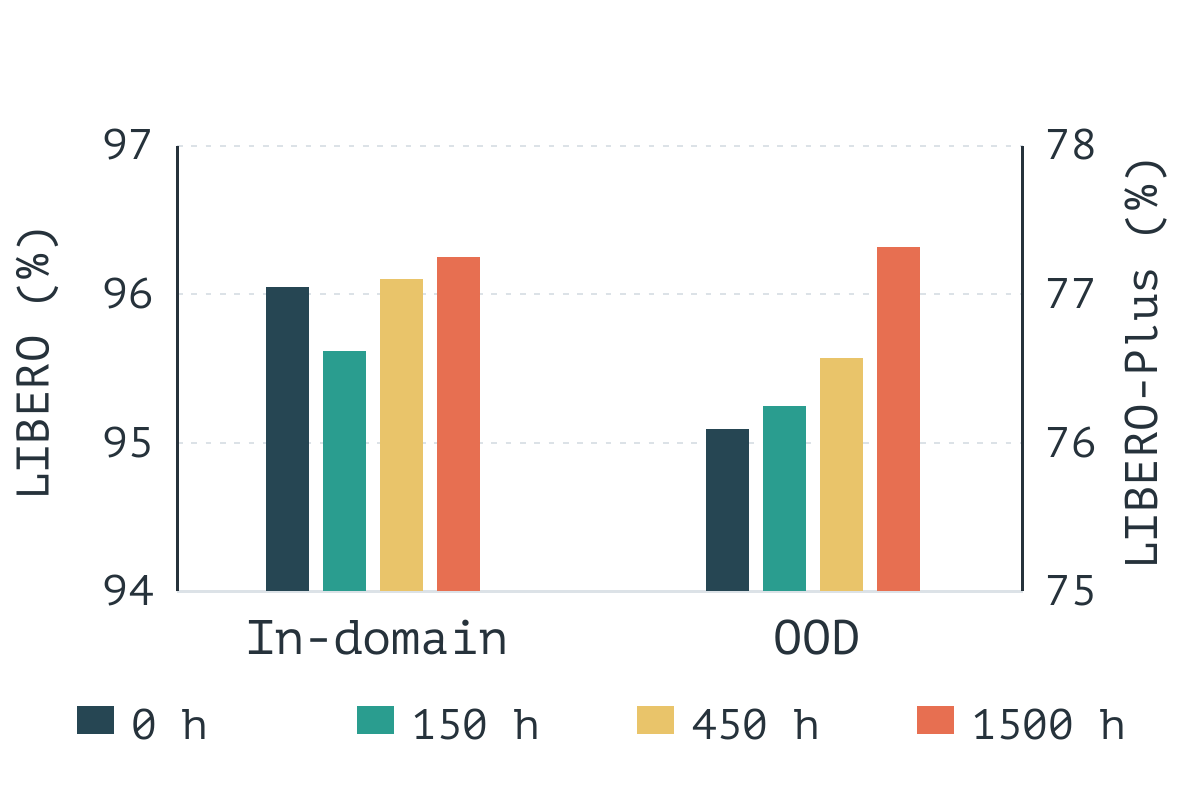}
    \caption{Video scale.}
    \label{fig:ego_data_scaling}
  \end{subfigure}\hfill
  \begin{subfigure}[b]{0.47\linewidth}
    \centering
    \includegraphics[width=\linewidth]{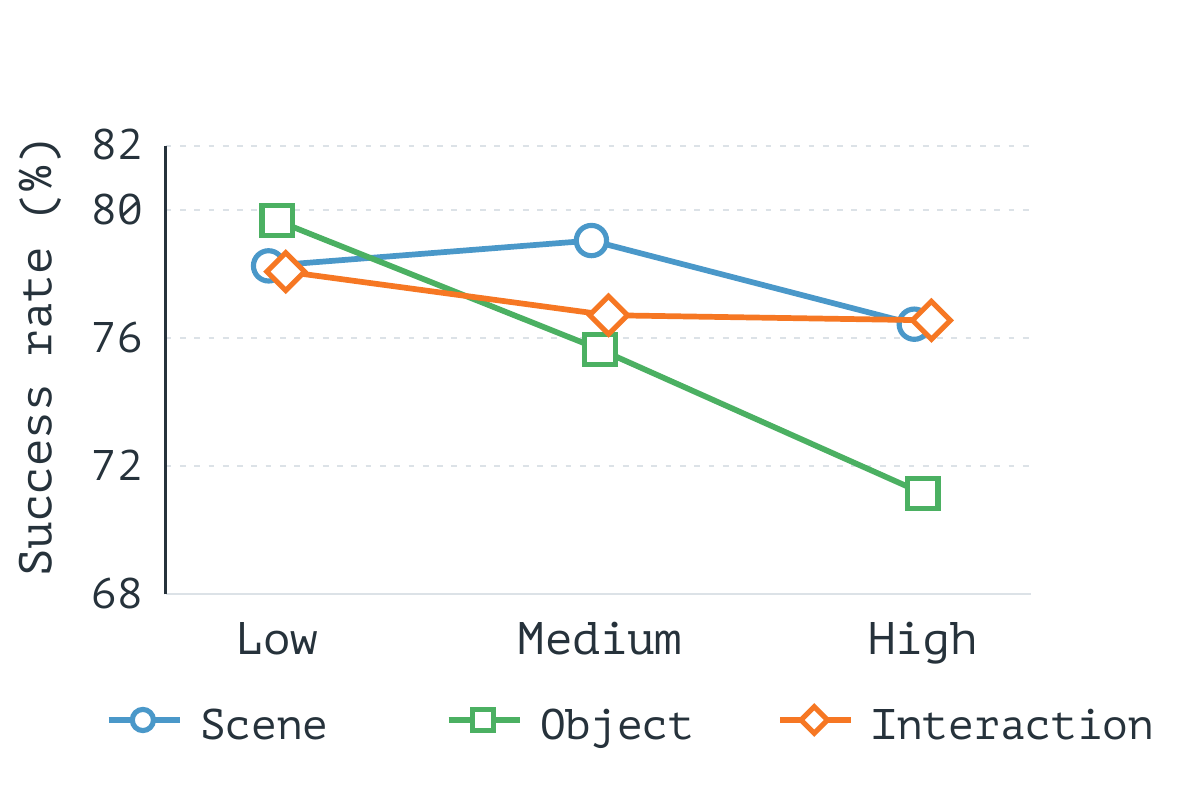}
    \caption{Diversity at a fixed budget.}
    \label{fig:ego_data_diversity}
  \end{subfigure}
  \caption{\textbf{Egocentric video data ablations.} (a) In-domain and OOD success with 0, 150, 450, and 1,500 hours of Open-AoE video co-training. (b) LIBERO-Plus success as scene, object, or interaction diversity varies within a fixed budget of 200K clip--description pairs.}
  \label{fig:data_ablate}
\end{figure}

\paragraph{Data Scaling.}
We then conduct a data scaling experiment with ego videos from 0 to 1,500 hours. We use the off-the-shelf Open-AoE, and its step description as the dataset source. As shown in Fig.~\ref{fig:ego_data_scaling}, In-domain LIBERO success changes little across these settings, as expected because the added human videos do not depict the downstream tasks. By contrast, LIBERO-Plus success rises steadily as video hours increase. Since these videos provide no action labels, the OOD trend is consistent with their visual prediction objective improving the shared VLM backbone and, in turn, the generalization of the downstream policy.

\paragraph{Data Diversity.}
We also study whether we can trade off some valid video duration for diversity. However, as shown in Fig.~\ref{fig:ego_data_diversity}, at a fixed egocentric video budget (200K clip-description pairs sampled from Open-AoE), increasing diversity along scene, object, or interaction dimensions does not yield a consistent gain in downstream success. Neither scene nor interaction diversity improves monotonically, while the highest object-diversity setting performs worse in this preliminary comparison. Together with the continuing gains from adding more hours, this suggests that egocentric video has not saturated the model over the tested range: expanding the amount of data across scenes, objects, and interactions appears more promising than maximizing any one diversity measure within a fixed video budget.

\section{Real-World Robot Evaluation}
\label{sec:real_robot}

We evaluate our model on a physical bimanual platform and verify: (1) Can egocentric pretraining enable our model to perform comparably to a strong VLA baseline on real-world manipulation tasks? (2) Does it retain this performance under visual perturbations unseen during fine-tuning?

\begin{figure}
  \centering
  \begin{subfigure}[b]{0.26\linewidth}
    \centering
    \includegraphics[width=\linewidth]{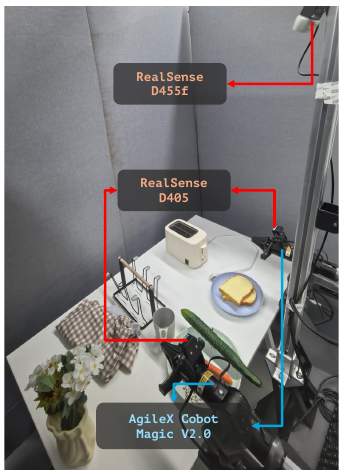}
    \label{fig:real_setup}
  \end{subfigure}\hfill
  \begin{subfigure}[b]{0.68\linewidth}
    \centering
    \includegraphics[width=\linewidth]{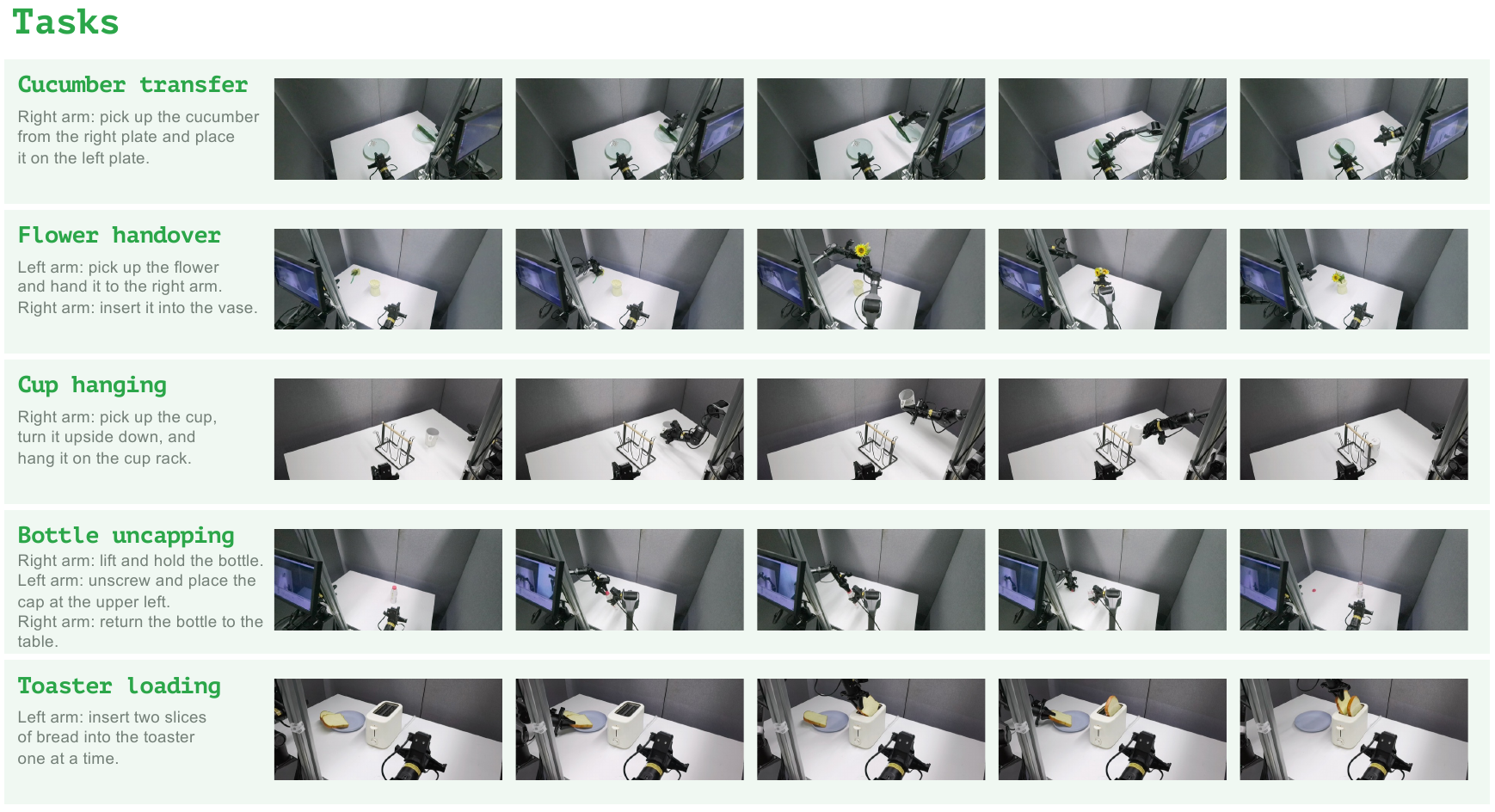}
    \label{fig:real_tasks}
  \end{subfigure}
  \caption{\textbf{Real-world evaluation setup.} Left: The hardware setup; objects from several tasks are placed together for illustration only. Right: Five sequential snapshots per task.}
  \label{fig:real_overview}
\end{figure}

\paragraph{Hardware.}
As shown in Fig.~\ref{fig:real_overview}, we use an AgileX Cobot Magic V2.0 with two 6-DoF arms and parallel-jaw grippers. We replace the stock wrist cameras with RealSense D405 cameras and add a top-mounted RealSense D455f. The policy takes images from all three cameras as input. The action space has 14 dimensions: six joint positions and one gripper command per arm.

\paragraph{Tasks.}
We design five tasks that vary in precision, inter-arm coordination, and horizon.
\emph{Cucumber Transfer}: The right arm moves a cucumber from the right plate to the left plate. This task serves as a basic pick-and-place test.
\emph{Flower Handover}: The left arm picks up a flower and hands it to the right arm in midair. The right arm then inserts the flower into a vase. This task tests coordinated positioning and grasp timing between the arms.
\emph{Cup Hanging}: The right arm grasps a cup by its handle, turns it upside down, and hangs it on a rack. The small clearance between the gripper and the handle opening requires precise positioning.
\emph{Bottle Uncapping}: The right arm lifts and holds a bottle. The left arm unscrews the cap and places it on the upper-left area of the table, and the right arm returns the bottle to the table. This task tests rotational manipulation and bimanual coordination.
\emph{Toaster Loading}: The left arm inserts two slices of bread into the toaster one at a time. Each insertion requires precise alignment, and the repeated pick-and-insert sequence extends the horizon.


\paragraph{Training and Evaluation.}
We collect demonstrations by tele-operation (Table~\ref{tab:real_main}). Bottle Uncapping has only 57 demonstrations and serves as a few-shot setting. We compare our model against $\pi_{0.5}$~\cite{black2025pi05}, a VLA pretrained on large-scale robot data and finetune it using the official \texttt{pi05\_base} weight. For each method and task, we fine-tune a separate checkpoint for 20k steps. Both models predict 50-step action chunks and execute 30 steps per chunk with asynchronous inference at 30\,Hz.

We evaluate each checkpoint for 10 trials per task. Initial object positions are randomized within a 4\,cm\,$\times$\,4\,cm region.
A trial is successful only if the robot completes the entire task without failure, as judged by a human evaluator. We report the per-task success rate and its average over the five tasks.

\begin{table}[t]
  \centering
  \caption{\textbf{Real-world Results}. \# Demos denotes the number of tele-operation demonstration collected for each task. We report the success rate out of 10 trials for each task.}
  \label{tab:real_main}
  \small
  \begin{tabular}{lcccccc}
    \toprule
     & Cucumber & Flower & Toaster & Cup & Bottle & \multirow{2}{*}{Total} \\
     & Transfer & Handover & Loading & Hanging & Uncapping & \\
    \midrule
    \# Demos    & 220 & 160 & 150 & 139 & 57 & -- \\
    \midrule
    $\pi_{0.5}$ & \textbf{9/10} & \textbf{9/10} & \textbf{2/10} & \textbf{1/10} & \textbf{1/10} & 22/50 \\
    Ours & \textbf{9/10} & 8/10 & \textbf{2/10} & 0/10 & \textbf{1/10} & 20/50 \\
    \bottomrule
  \end{tabular}
\end{table}

\paragraph{Results.}
As shown in Table~\ref{tab:real_main}, our model achieves an average success rate of 40\%, compared with 44\% for $\pi_{0.5}$. The two methods tie on three tasks and differ by one trial on the other two. Notably, our model performs comparably to $\pi_{0.5}$ despite using a smaller robot pretraining corpus and training its action expert from scratch. This suggests that the visual-state prediction on human videos provides transferable priors that partly compensate for both.
Both methods achieve high success on Cucumber Transfer and Flower Handover but rarely complete Cup Hanging, Bottle Uncapping, or Toaster Loading. Precision grasping, bimanual rotational manipulation, and long-horizon execution remain challenging for both methods. Most failures of our model occur during grasping, which is likely related to its smaller robot pretraining corpus.

\begin{figure}[t]
  \centering
  \includegraphics[width=\linewidth]{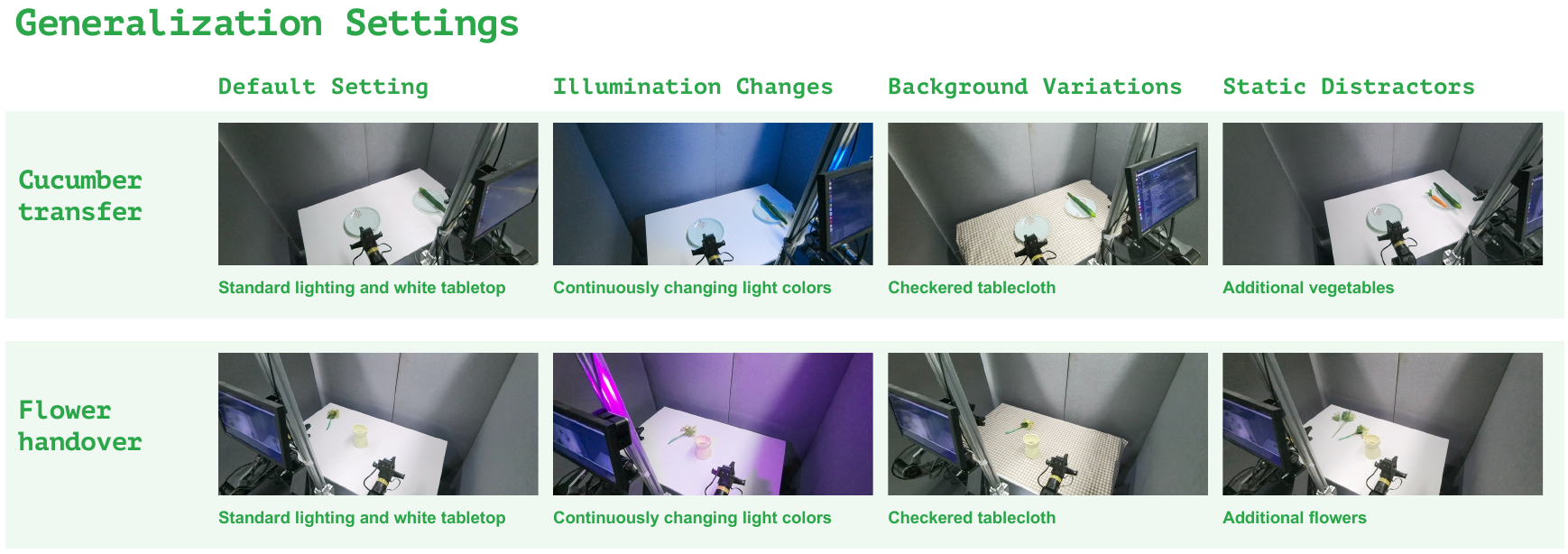}
  \caption{\textbf{Real-world generalization} evaluation setting under visual perturbations.}
  \label{fig:real_perturb}
\end{figure}

\paragraph{Visual perturbations.}
We apply three perturbations to Cucumber Transfer and Flower Handover (Fig.~\ref{fig:real_perturb}). For \emph{illumination changes}, we dim the white light and turn on a colored light that cycles at 1.5\,Hz. For \emph{background variations}, we cover the tabletop with a checkered tablecloth. For \emph{static distractors}, we place a similar object next to the target: a carrot beside the cucumber and additional flowers beside the flower. We evaluate each perturbation separately for 10 trials per task.

\begin{table}[t]
  \centering
  \caption{\textbf{Success under visual perturbations} out of 10 trials for each task. Default: unperturbed setting; Illum.: illumination changes; Backgr.: background variations; Distr.: static distractors.}
  \label{tab:real_perturb}
  \small
  \setlength{\tabcolsep}{4.5pt}
  \begin{tabular}{lcccccccc}
    \toprule
     & \multicolumn{4}{c}{Cucumber Transfer} & \multicolumn{4}{c}{Flower Handover} \\
    \cmidrule(lr){2-5}\cmidrule(lr){6-9}
    Method & Default & Illum. & Backgr. & Distr. & Default & Illum. & Backgr. & Distr. \\
    \midrule
    $\pi_{0.5}$ & \textbf{9} & 7 & 7 & 4 & \textbf{9} & \textbf{5} & \textbf{2} & \textbf{3} \\
    Ours       & \textbf{9} & \textbf{9} & \textbf{8} & \textbf{5} & 8 & 2 & \textbf{2} & 0 \\
    \bottomrule
  \end{tabular}
\end{table}

\paragraph{Results under Perturbations.}
Table~\ref{tab:real_perturb} reports the results. Averaged over both tasks, static distractors cause the largest drop for both methods, from 90\% to 35\% for $\pi_{0.5}$ and from 85\% to 25\% for ours.
On Cucumber Transfer, both methods achieve 9/10 in the default setting. Our model outperforms $\pi_{0.5}$ under all three perturbations. Under illumination changes, our model maintains 9/10, whereas $\pi_{0.5}$ drops to 7/10. Under background variations and static distractors, our model achieves 8/10 and 5/10, compared with 7/10 and 4/10 for $\pi_{0.5}$. Consistent with the LIBERO-Plus results (Sec.~\ref{sec:ego_data}), this suggests that visual-state prediction on diverse human videos yields representations that are less sensitive to appearance changes.
On Flower Handover, our model matches $\pi_{0.5}$ under background variations (2/10) but underperforms it under illumination changes (2/10 vs.\ 5/10) and static distractors (0/10 vs.\ 3/10). We hypothesize that $\pi_{0.5}$ benefits here from action priors learned from large-scale bimanual robot data.

\section{Conclusion}
We investigated whether human egocentric videos without action labels can benefit robot policy learning through future visual-state prediction. During model design exploration, we find that learned visual representations are more effective prediction targets than reconstruction latents under distribution shift, and regression is a viable alternative to flow matching under the Shared Prediction on Action and Representation (SPAR) design. Data quality also matters: filtering Ego4D for stable motion and visible hands improves on both raw-video co-training and the robot-only baseline. As the amount of action-free video increases, OOD success rises consistently. On real-world manipulation tasks, our policy performs comparably to the powerful VLA baseline, with task-dependent results under visual perturbations. Together, these findings provide fresh insight for the utilization of action-free egocentric video, showing its potential as a useful source of transferable visual supervision.
{
\small
\bibliographystyle{unsrtnat}
\bibliography{ref}
}

\appendix
\section{Implementation Details}
\label{app:impl}
 
\paragraph{Backbone and inputs.}
We use Qwen3.5-2B~\citep{qwen35}, whose LLM has $L{=}24$ layers of width $D{=}2048$. Observations are processed at native resolution with a budget of $256{\times}256$ pixels per image. Each image is preceded by a description of its camera role: ``primary global view from a fixed exterior camera'' and ``local effector view from a wrist camera'' for DROID, ``primary global view from a fixed front camera'' and ``local effector view from a wrist camera'' for LIBERO, and ``primary global view from a moving human head camera'' for egocentric video. For egocentric clips, the automatically generated step description serves as the instruction (App.~\ref{app:ego4d_curation}). The generation target is always the future frame of the primary global view, i.e., the third-person camera for robots and the head camera for humans. The vision encoder, including its token-merging projector, is frozen; the LLM and both experts are trained.
 
\paragraph{Expert architecture.}
Both experts have $L$ blocks of width $D$ ($32$ heads of dimension $64$, dropout $0.2$), and block $l$ is paired with LLM layer $l$. Blocks alternate between cross-attention to the paired hidden states $\mathbf{h}^{(l)}$ and self-attention among the expert tokens, each modulated by AdaLN~\citep{peebles2023dit} on a conditioning embedding $\mathbf{c}$:
\begin{equation}
  \tilde{\mathbf{x}}=\mathbf{x}+\operatorname{Attn}\big(\operatorname{AdaLN}(\mathbf{x};\mathbf{c}),\,\mathbf{m}^{(l)}\big),\qquad
  \mathbf{x}\leftarrow\tilde{\mathbf{x}}+\operatorname{FFN}\big(\operatorname{LN}(\tilde{\mathbf{x}})\big),
\end{equation}
where $\mathbf{m}^{(l)}=\mathbf{h}^{(l)}$ in cross-attention blocks and the normalized expert tokens otherwise. The two experts never attend to each other.
\begin{itemize}
  \item \emph{Action expert.} The input is $32$ learnable tokens followed by $H$ action tokens, each embedding one row of the noised chunk together with a sinusoidal encoding of the flow time and a learned positional embedding; a proprioceptive-state token can optionally be prepended (not used in our experiments). The blocks are conditioned on $\mathbf{c}=\operatorname{emb}(s)$, and an MLP maps the outputs at the action positions to a velocity.
  \item \emph{Regression generation expert.} The input is $N$ learnable query tokens with positional embeddings, one per target token. The conditioning is constant, $\mathbf{c}=\operatorname{emb}(0)$, and a three-layer MLP with an output LayerNorm maps the outputs to $\hat{\mathbf{Z}}_t\in\mathbb{R}^{N\times d}$.
  \item \emph{Diffusion generation expert.} The noised target tokens are linearly projected to width $D$ and given learned positional embeddings; the blocks are conditioned on $\mathbf{c}=\operatorname{emb}(s)$, and an AdaLN-modulated output projection predicts the velocity in $\mathbb{R}^{N\times d}$.
\end{itemize}
We attach the generation branch as a separate expert rather than appending learnable generation queries to the LLM sequence: the latter changes the input on which the action pathway conditions, and in preliminary experiments it consistently yielded lower success rates.
 
\paragraph{Temporal alignment.}
The future frame $\mathbf{o}_{t+\Delta}$ is aligned with the end of the action chunk, $a_{t+H}$, so the visual prediction horizon in physical time is $\tau=H/f_{\mathrm{R}}$, where $f_{\mathrm{R}}$ is the frame rate of the robot data. It is converted to each stream's frame rate so that all streams predict the same instant: a stream recorded at $f_s$ frames per second uses an offset of $\Delta_s=\lfloor\tau f_s+\tfrac12\rfloor$ frames. During pretraining on DROID ($15$\,Hz), $\tau{=}8/15{\approx}0.53$\,s, i.e., $8$ DROID frames and $16$ frames of a $30$\,fps egocentric video; during fine-tuning on LIBERO ($20$\,Hz), $\tau{=}0.4$\,s, i.e., $8$ frames.
 
\paragraph{Generation targets.}
Tab.~\ref{tab:app_targets} lists the target encoders. All are frozen and applied online to the raw future frame, so switching targets requires no data re-processing. For the regression loss, each target is standardized as $\bar{\mathbf{Z}}_t=(\mathbf{Z}_t-\mu_\phi)/\sigma_\phi$ with scalar statistics estimated once on future frames of the training mixture.
 
\begin{table}[h]
  \centering
  \small
  \caption{Target encoders. $N$ and $d$ are the number and dimension of target tokens per frame.}
  \label{tab:app_targets}
  \setlength{\tabcolsep}{4.5pt}
  \begin{tabular}{@{}lllccc@{}}
    \toprule
    Target & Encoder & Pretraining objective & Res. & $N$ & $d$ \\
    \midrule
    Qwen-ViT & Qwen3.5-2B vision encoder~\citep{qwen35} & vision--language (VLM) & 224 & 49 & 2048 \\
    SigLIP2 & So400m/16~\citep{tschannen2025siglip2} & language--image & 256 & 256 & 1152 \\
    DINOv2 & ViT-g/14~\citep{oquab2024dinov2} & self-distillation & 224 & 257 & 1536 \\
    LingBot & LingBot-Vision ViT-g/16~\citep{lingbotvision} & masked boundary modeling & 512 & 1024 & 1536 \\
    V-JEPA2\textsuperscript{$\dagger$} & ViT-g/16~\citep{assran2025vjepa2} & latent video prediction & 256 & 256 & 1408 \\
    Wan-VAE & Wan2.2 VAE~\citep{wan2025} & pixel reconstruction & 256 & 256 & 48 \\
    \bottomrule
  \end{tabular}\\[2pt]
  {\footnotesize \textsuperscript{$\dagger$}A video encoder; the single future frame is repeated to fill one temporal tubelet.}
\end{table}
 
\paragraph{Flow-matching schedules.}
For actions, following \cite{pi0}, we sample $s=1-u/0.999$ with $u\sim\mathrm{Beta}(1.5,1)$ and draw two $(s,\boldsymbol{\epsilon})$ pairs per example while reusing the VLM features. For the diffusion generation expert, the schedule depends on the target type. For representation targets, we follow RAE~\citep{zheng2025rae}: the noise level $\sigma=1-s$ is drawn from a logit-normal distribution~\citep{esser2024sd3} and shifted as $\sigma\leftarrow\gamma\sigma/(1+(\gamma-1)\sigma)$ with $\gamma=\max\big(1,\sqrt{Nd/4096}\big)$. For VAE latents, we follow common practice in video latent diffusion~\citep{wan2025}: $\sigma$ is drawn uniformly and shifted with $\gamma{=}5$, and the per-sample loss is weighted by $w(s)\propto\exp\big(-2(s-\tfrac12)^2\big)$, normalized to unit mean within a batch. Diffusion targets are not standardized.
 
\paragraph{Co-training.}
The per-GPU batch is split between the robot and egocentric streams by a sampling ratio ($1{:}1$ by default). The two mini-batches are forwarded separately, each computing only its bound losses, and their gradients are accumulated into a single optimizer step (Alg.~\ref{alg:cotrain}). Because each loss is averaged within its own mini-batch, the ratio changes how many samples of each source are seen per step but not the relative weight of the two objectives.
 
\begin{algorithm}[h]
  \caption{One co-training step}
  \label{alg:cotrain}
  \begin{algorithmic}[1]
    \Require robot mini-batch $\mathcal{B}_{\mathrm{R}}$; egocentric mini-batch $\mathcal{B}_{\mathrm{E}}$
    \For{$\mathcal{B}\in\{\mathcal{B}_{\mathrm{R}},\mathcal{B}_{\mathrm{E}}\}$}
      \State $\mathcal{H}\gets\mathrm{LLM}_\theta(\text{images},\text{role prompts},\ell)$ \Comment{per-layer hidden states}
      \State $\mathbf{Z}_t\gets\phi(\mathbf{o}_{t+\Delta})$ \Comment{frozen encoder, no gradient}
      \State $\mathcal{L}_{\mathcal{B}}\gets\lambda\,\mathcal{L}_{\text{gen}}(g_\omega;\mathcal{H},\mathbf{Z}_t)$
      \If{$\mathcal{B}=\mathcal{B}_{\mathrm{R}}$}
        \State $\mathcal{L}_{\mathcal{B}}\gets\mathcal{L}_{\mathcal{B}}+\mathcal{L}_{\text{action}}(f_\psi;\mathcal{H},\mathbf{A}_t)$
      \EndIf
      \State accumulate $\nabla_{\theta,\psi,\omega}\mathcal{L}_{\mathcal{B}}$
    \EndFor
    \State update $(\theta,\psi,\omega)$ with AdamW
  \end{algorithmic}
\end{algorithm}
 
\paragraph{Optimization.}
Tab.~\ref{tab:app_hparams} lists the training hyperparameters. We train with bfloat16 mixed precision, DeepSpeed ZeRO-2, and gradient checkpointing.
 
\begin{table}[h]
  \centering
  \small
  \caption{Training hyperparameters.}
  \label{tab:app_hparams}
  \begin{tabular}{@{}ll@{}}
    \toprule
    Optimizer & AdamW, $\beta=(0.9,0.95)$, $\epsilon=10^{-8}$, weight decay $10^{-8}$ \\
    Learning rate & LLM $2.83\times10^{-5}$; experts $2.83\times10^{-4}$; other $1.41\times10^{-4}$ \\
    Schedule & 500 warmup steps, cosine decay to $2.83\times10^{-6}$ \\
    Pretraining steps & 20{,}000 \\
    Downstream fine-tuning steps & 20{,}000 \\
    Pretraining global batch size & 1{,}024 (1:1 robot/egocentric) \\
    Gradient clipping & 1.0 \\
    Action chunk & $H{=}8$, $d_a{=}7$ \\
    Loss weights & $\lambda_{\text{action}}=\lambda_{\text{gen}}=1$ \\
    Future frames & 1 \\
    Inference & 4 Euler steps \\
    \bottomrule
  \end{tabular}
\end{table}
 
\paragraph{Inference.}
Starting from Gaussian noise, the action chunk is obtained with $M{=}4$ Euler steps, $\mathbf{A}^{s_{k+1}}_t=\mathbf{A}^{s_k}_t+\tfrac{1}{M}f_\psi(\mathbf{A}^{s_k}_t,s_k;\mathcal{H})$ with $s_k=k/M$. The generation expert is not executed.

\section{Ego4D Data Curation}
\label{app:ego4d_curation}

\paragraph{Clip construction.}
We partition each source video into uniform, non-overlapping 5-second clips. The two quality gates are applied in sequence to sampled points within each clip, following the EgoSmith procedure~\citep{zhong2026EgosteerorEgosmith}.

\paragraph{Gate 1: motion quality.}
Corner detection and optical-flow fitting provide a point-wise motion check. Let $N_F$ (\texttt{flow\_evaluable\_count}) be the number of flow-evaluable sampled points in a clip, excluding the first sampled point of a source video when no preceding frame is available. Let $n_F$ (\texttt{flow\_pass\_count}) count those points for which the optical-flow displacement check passes (\texttt{motion\_ok=True}), and let $L_F$ (\texttt{flow\_max\_invalid\_run}) be the longest consecutive run of failed checks (\texttt{motion\_ok=False}) among flow-evaluable points. Define $r_F=n_F/N_F$ (\texttt{flow\_pass\_ratio}) when $N_F>0$. The clip passes Gate 1 only if all three conditions hold:
\begin{equation}
  r_F\geq 0.85,\qquad n_F\geq 3,\qquad L_F\leq 2.
\end{equation}
A clip with $N_F=0$ has an undefined pass ratio and fails the gate. This gate favors limited camera motion with reliable optical-flow estimates.

\paragraph{Gate 2: visible hands.}
A YOLO hand detector is applied to every sampled frame. A detected hand bounding box is valid only if its confidence is at least $0.30$, its area is between $2\%$ and $50\%$ of the image area, and it intersects the normalized image region $[0.075,0.075,0.925,1.0]$ (left, top, right, bottom). Let $h_i$ (\texttt{hand\_count}) be the number of valid hand boxes in sampled frame $i$. For a clip with $N$ sampled frames, let $n_H=\sum_{i=1}^{N}\mathbf{1}[h_i\geq1]$ (\texttt{hand1\_pass\_count}), let $r_H=n_H/N$ (\texttt{hand1\_pass\_ratio}), and let $L_H$ (\texttt{hand1\_max\_invalid\_run}) be the longest consecutive run of frames with $h_i<1$. The clip passes Gate 2 only if
\begin{equation}
  r_H\geq 0.10,\qquad n_H\geq 1,\qquad L_H\leq 5.
\end{equation}
Clips with no sampled frames fail this gate. This criterion requires at least one sufficiently visible hand; it does not require two hands.

\paragraph{Caption generation.}
For each 5-second clip, we sample every tenth frame to form a sequence of 15 frames and provide it to a locally deployed Qwen3.6-35B-A3B model. We require a non-empty English \texttt{step\_description}: one imperative sentence beginning with a verb and describing the principal visible activity. The field must remain non-empty even if the clip is marked \texttt{Invalid}; in that case, the description should conservatively state the visible non-manipulation activity without inventing an object interaction. The resulting description is used as the text instruction when predicting future visual states from the clip.

\end{document}